# Revisiting Linformer with a modified self-attention with linear complexity

Madhusudan Verma

Independent Researcher

vermamadhusudan2020@gmail.com

**Abstract**

*Although Transformer models such as Google's BERT and OpenAI's GPT-3 are successful in many natural language processing tasks, training and deploying these models are costly and inefficient.Even if pre-trained models are used, deploying these models still remained a challenge due to their large size. Apart from deployment, these models take higher time during inference restricting user-friendliness. The main bottleneck is self-attention which uses quadratic time and space with respect to the sequence length. In order to reduce the quadratic time complexity of the self-attention mechanism, Linformer by Facebook's AI research team was introduced where they showed that the self-attention mechanism can be approximated by a low-rank matrix and exploiting this finding, a new method for self-attention with linear time and space complexity was proposed by them. In the Linformer ,the time complexity depends on the projection mapping dimension which acts as a hyperparameter and affects the performance of the model, tuning this hyperparameter can be time-consuming. In this paper, I proposed an alternative method for self-attention with linear complexity in time and space and is independent of the projection mapping dimension. Since this method works for long sequences this can be used for images as well as audios.*

## 1. Introduction and related work

Transformer(Vaswani et al 2017) has become the popular model for natural language processing including text classification, translation(Ott et al., 2018), or question answering system. Models that uses transformer have a huge number of parameters starting from 340 million in BERT-large to 175 billion in GPT-3. Due to this training and deploying such models are slow and require extensive distillation or compression to use for real-life applications.

The main bottleneck is self-attention which requires $O(n^2)$ There were prior works done to reduce this complexity One method was to introduce sparsity into the attention layers by making each token to attend only a subset of tokens of an entire sequence. But this method suffers from a large performance drop with limited efficiency gain. Then Reformer was introduced which uses locally sensitive hashing was used to avoid costly computation they also proposed to use reversible layers to allow storing the only once instead of for each layer but its efficiency gain appears only after sequence length >;2048.

| Model | Complexity per layer | Sequential Operation |
|---|---|---|
| Recurrent Neural Network | $O(n)$ | $O(n)$ |
| Transformer | $O(n^2)$ | $O(1)$ |
| Sparse Transformer | $O(n\sqrt{n})$ | $O(1)$ |
| Reformer | $O(n \log(n))$ | $O(\log(n))$ |
| Linformer | $O(nk)$ | $O(1)$ |
| This model | $O(nd^2)$ | $O(1)$ |

**Table 1:** Per-layer time complexity and minimum number of sequential operations as a function of sequence length (n) for various architectures, k is the projection dimension and d is the embedding dimension

## 2. Background

Let Q,K,V be the key, query and value the attention is defined as

$$soft\max\left(\frac{QW_i^Q\left(KW_i^K\right)^T}{\sqrt{d_k}}\right)VW_i^V$$

Where $W_i^Q$ and $W_i^K$ are the matrices learned during training

If we use dot product and use a general function then the attention becomes

$$f\left(QW_i^Q\left(KW_i^K\right)^T\right)VW_i^V$$

Now if f is scaling function $f(x)=\frac{x}{n}$

Then above equation becomes $\left(\frac{QW_i^Q\left(KW_i^K\right)^T}{n}\right)VW_i^V$

If we take $f\left(QW_i^Q\right)f\left(KW_i^K\right)^T VW_i^V$ as attention and $f(x)=\frac{x}{\sqrt{n}}$ as scaling function then both are equivalent

$$\frac{QW_i^Q}{\sqrt{n}}\left(\frac{(KW_i^K)^T}{\sqrt{n}}VW_i^V\right)$$

$$\overset{(a)}{=} \frac{QW_i^Q}{n}\left((KW_i^K)^T VW_i^V\right)$$

$$= \left(\frac{QW_i^Q(KW_i^K)^T}{n}\right)VW_i^V$$

Step a utilizes the fact that scalar multiplication is commutative and matrix multiplication is associative.

**JL lemma:**
For a given $0 < \varepsilon < 1$ and a set $Y$ of q points in $R^n$ such that $n > 8\ln(q)/\varepsilon^2$ there exists a linear map f: $R^N \to R^n$ such that
$$(1-\varepsilon)\|x-y\|^2 \leq \|f(x)-f(y)\|^2 \leq (1+\varepsilon)\|x-y\|^2$$
for all $x, y \in Y$

**Compact region:**
Let's first look at the definition of compact set for real numbers. A set S of real numbers is called compact if every sequence in S has a subsequence that converges to an element again contained in S. Formally, a topological space $X$ is called compact if each of its open covers has a finite subcover

**Lipchitz continuous functions:**
Given two metric spaces $(X, d_X)$ and $(Y, d_Y)$, where $d_X$ denotes the metric on the set X and $d_Y$ is the metric on set Y, a function f : X → Y is called Lipschitz continuous if there exists a real constant K ≥ 0 such that, for all $x_1$ and $x_2$ in X,

$$d_Y(f(x_1), f(x_2)) \leq K d_X(x_1, x_2)$$

## 3. Method

### 3.1 Proof that the new context mapping matrix is low rank

The following proof is based on JL Lemma((Lindenstrauss, 1984) the following version is from (Arriaga & Vempala, 2006).

Let R be an $k \times n$ matrix with i.i.d entries from $N\left(0, \frac{1}{k}\right)$. For any $x, y \in R^n$ we have

**Lemma1**
$$\Pr(\|Rx\| \leq (1+\varepsilon)\|x\|) \geq 1 - 2e^{-(\varepsilon^2-\varepsilon^3)k/4}$$

**Lemma2**
$$\Pr(\|xR^T Ry^T\| \leq \varepsilon\|xy\|) \geq 1 - 2e^{-(\varepsilon^2-\varepsilon^3)k/4}$$

Let $A = \dfrac{QW_i^Q}{\sqrt{d_k}}$, $B = \dfrac{(KW_i^K)^T}{\sqrt{d_k}}$

Then,

$$P = \text{soft}\max(A)\text{soft}\max(B)$$
$$= \exp(A)D_A^{-1}\exp(B)D_B^{-1}$$

Where $D_A$ and $D_B$ are diagonal matrices

$$(D_A)_{ii} = \sum_{j=1}^{n}\exp(A_{ji}) \text{ and } (D_B)_{ii} = \sum_{j=1}^{n}\exp(B_{ji})$$

Let's define

$$\tilde{P} = \exp(A)D_A^{-1}\exp(B)D_B^{-1}R^T R$$

We know that for any two matrices compatible for multiplication

$$\text{rank}(LM) \leq \min\{\text{rank}(L), \text{rank}(M)\}$$

$$\text{rank}\left(\exp(A)D_A^{-1}\exp(B)D_B^{-1}R^T R\right) \leq \{\text{rank}\left(\exp(A)D_A^{-1}\exp(B)D_B^{-1}R^T\right), \text{rank}(R)\} \leq \text{rank}(R) = k$$

Now we will prove that for any column vector $c \in R^n$

$$\Pr\left(\left\|\tilde{P}c - Pc\right\| \leq \varepsilon\|Pc\|\right) > 1 - o(1)$$

for some $k$

Applying lemma2 for any row vector $u \in R^n$ of P matrix and for any column vector $w \in R^n$ of $VW_i^V$ we obtain

$$\Pr\left(\left\|uR^T Rw^T - uw^T\right\| \leq \|uw^T\|\right) > 1 - 2e^{-(\varepsilon^2 - \varepsilon^3)k/4} \qquad (1)$$

Therefore we have

$$\Pr\left(\left\|\tilde{P}w^T - Pw^T\right\| \leq \|Pw^T\|\right) = \Pr\left(\left\|\tilde{P}R^T Rw^T - Pw^T\right\| \leq \|Pw^T\|\right)$$

$$\geq 1 - \sum_{x \in P}\left(\left\|xR^T Rw^T - Pw^T\right\| \leq \|xw^T\|\right)$$

$$> 1 - 2ne^{-(\varepsilon^2 - \varepsilon^3)k/4}$$

Where step (a) is based on union bound and (b) is using equation

setting $k = 5\log(n)/(\varepsilon^2 - \varepsilon^3)$ the result follows
Here is the demonstration for the same in figure 1

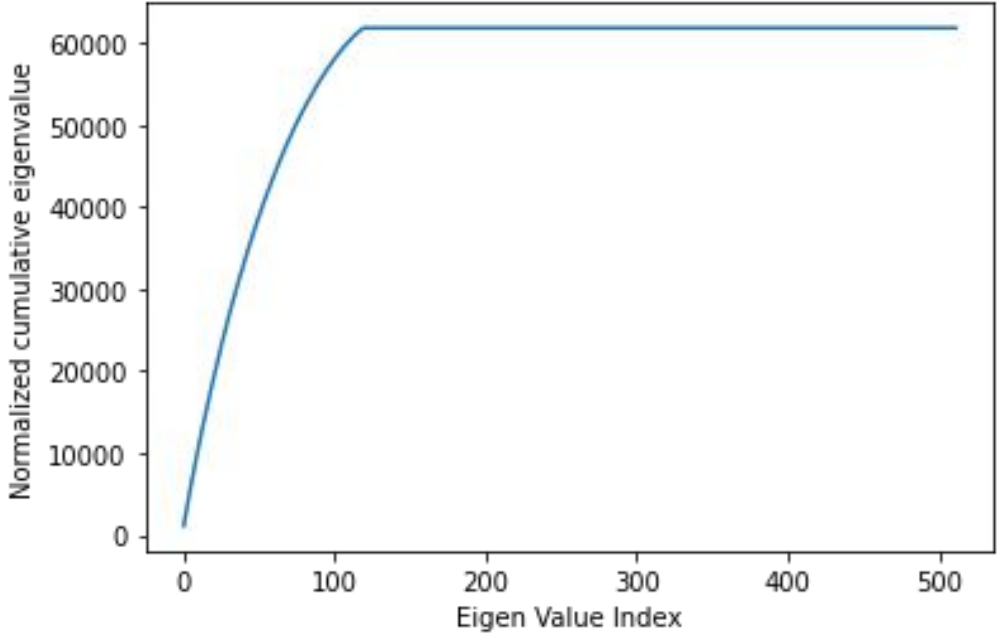

**Figure1:** The long tail indicates that the matrix can be approximated by a low rank matrix

## 3.2 Method to contruct low-rank matrix and it is close to new context mapping matrix

Now we define a new method to find attention,

$$head_i = Attention(QW_i^Q, E_i KW_i^K, F_i VW_i^V)$$

$$= soft\max\left(\frac{QW_i^Q}{\sqrt{d_k}}\right) soft\max\left(\frac{(E_i KW_i^K)^T}{\sqrt{d_k}}\right) F_i VW_i^V$$

Let $H = Soft\max\left(\frac{QW_i^Q}{\sqrt{d_k}}\right)$, $L = soft\max\left(\frac{(E_i KW_i^K)^T}{\sqrt{d_k}}\right)$ and $X = F_i VW_i^V$

Size of H is n×d, L is d×k and X is k×d according to the paper for Linformer
Since Matrix multiplication is associative
Time complexity of (HL)X = ndk+ndk=2ndk= $O(ndk)$
Time complexity of H(LX)=ndd+dkd= $O(nd^2)$
If we multiply in the second way we can see it is independent of k.

Now we need to prove that for $x_1 \in H$ and $x_2 \in L$

$$\Pr\left(\left\|\exp(x_1)\exp(x_2 E_i)F_i V W_i^V - \exp(x_1)\exp(x_2)V W_i^V\right\| \leq \varepsilon \left\|\exp(x_1)\exp(x_2)V W_i^V\right\|\right) \geq 1 - o(1)$$

This can be proved by proving for any $y \in V W_i^V$

$$\Pr\left(\left\|\exp(x_1)\exp(x_2 E_i)F_i y^T - \exp(x_1)\exp(x_2)y^T\right\| \leq \varepsilon \left\|\exp(x_1)\exp(x_2)y^T\right\|\right) \geq 1 - 2e^{-(\varepsilon^2-\varepsilon^3)k/4}$$
(2)

Since for two vectors a and b $\|ab\| = \|a\|\|b\|$
Equation (1) can be rewritten as

$$\Pr\left(\left\|\exp(x_2 E_i)F_i y^T - \exp(x_2)y^T\right\| \leq \varepsilon \left\|\exp(x_2)y^T\right\|\right) \geq 1 - 2e^{-(\varepsilon^2-\varepsilon^3)k/4}$$
(3)

Let $E_i = \delta R$ and $F_i = e^{-\delta} R$ where $R \in R^{k \times n}$ with i.i.d entries from $N\left(0, \frac{1}{k}\right)$ and $\delta = 1/2^n$

By Triangle inequality of norms, we have the following

$$\left\|\exp(x_2 E_i)F_i y^T - \exp(x_2)y^T\right\| \leq \varepsilon\|\exp(x_2)\| \leq \left\|\exp(x_2 E_i)F_i y^T - \exp(x_2 R^T R y)\right\| + \left\|\exp(x_2 R^T R y) - \exp(x_2)y^T\right\|$$

$$\overset{(a)}{\leq} (1+\varepsilon)\|y\|\|\exp(x_2 E_i) - \exp(x_2)\| + \left\|\exp(x_2 R^T R y) - \exp(x_2)y^T\right\|$$

$$\overset{(b)}{\leq} \left\|\exp(x_2 R^T R y) - \exp(x_2)y^T\right\| + o(\|\exp(x_2)\|\|y\|)$$

$$\overset{(c)}{\leq} \varepsilon(\|\exp(x_2)\|\|y\|) + o(\|\exp(x_2)\|\|y\|)$$

(a) is based on cauchy inequality and lemma 1 ( b) is based on the fact that exponential function is lipchitz continuous in compact region then we can choose small enough $\delta = o\left(\frac{1}{n}\right)$ such that $\|\exp(\delta x R) - \exp(\delta x)R\| = o(\|\exp(x_2)\|)$
And c is by lemma 2

Setting $k = 5\log(n)(\varepsilon^2 - \varepsilon^3)$ the result follows

Note that k is dependent on sequence length n, but it can be proved for k independent of n using the fact that P is low rank

Here is the proof for the same
Let rank(P)=d, then we can find a submatrix $P^s \in R^{2d \times d}$ of $\exp(A)\exp(BE^T)FV$ such that rank of this submatrix is also d.

Applying the result in equation - for every row of matrix $P^s$ and every column of V $k = 9\log(d)(\varepsilon^2 - \varepsilon^3)$ and we obtain for any row $P_i^s$ of $P^s$

$$\Pr\left(\left\|\exp(P_i^s E_i)F_i V - \exp(P_i^s)V\right\| \leq \varepsilon \left\|\exp(P_i^s)V\right\|\right) \geq 1 - o(1) \quad (4)$$

P=AB in a similar way we can write $P^s = A'B'$
Let's define a matrix

$$\Gamma = \begin{bmatrix} \exp(A)\exp(BE^T)FV \\ \exp(A)\exp(B)V \end{bmatrix} \begin{bmatrix} \exp(A')\exp(B'E^T)FV \\ \exp(A')\exp(B')V \end{bmatrix}^{-1}$$

Size of the above matrix is $n \times 2d$
Then, for every row $B_i$ of B $P_i^s$ of $P^s$ and $\Gamma_i$ of $\Gamma$
$$\left\|\exp(B_i E^T)FV - \exp(B_i)V\right\| = \left\|\Gamma_i \exp(P^s E^T)FV - \Gamma_i \exp(P^s)V\right\|$$
$$\overset{(a)}{\leq} \left\|\exp(P^s E^T)FV - \exp(P^s)V\right\|_2 \left\|\Gamma_i\right\|$$
$$\overset{(b)}{\leq} \Theta(d) \sum_{i=1}^{2d} \left\|\exp(P^s E^T)FV - \exp(P^s)V\right\|_F$$
$$= \Theta(d) \sum_{i=1}^{2d} \left\|\exp(P_i^s E^T)FV - \exp(P_i^s)V\right\|$$
$$\overset{(c)}{\leq} \varepsilon \Theta(d) \sum_{i=1}^{2d} \left\|\exp(P_i^s)\right\| \|V\|$$
$$\leq \varepsilon \Theta(d) \left\|\exp(P_i^s)\right\| \|V\|$$

Step (a) is from the fact that $\|Ax\| \leq \|A\|_2 \|x\|$ where $\|A\|_2 = \sqrt{\lambda_{\max} A^T A}$ Step (b) uses matrix norm inequality $\|A\|_2 \leq \|A\|_F$. (c) is obtained by using equation (4).

# 4. Conclusion

In this paper, I introduced a new method for calculating self-attention. I showed theoretically and empirically that there exists a low-rank matrix for this matrix also. I then proposed a method to calculate this low-rank matrix and proved theoretically that the project matrix is not far from the original matrix.